\documentclass[final,1p,times,twocolumn,authoryear]{elsarticle}

\usepackage{amssymb}
\usepackage{amsmath}
\usepackage{hyperref}
\usepackage{float}

\begin{document}

\begin{frontmatter}


\title{OSSDD - a New Open Dataset for Sentinel-1 Ship Detection} 

\author[label1]{Horst Hammer \corref{cor1}}
\ead{horst.hammer@iosb.fraunhofer.de}
\author{Sylvia Hochstuhl\texorpdfstring{$^{\textup{a,b}}$}{}}
\author{Antje Thiele\texorpdfstring{$^{\textup{a,b}}$}{}}
\author[label3]{Tobias Brosch}
\author[label3]{Padraig Davidson}
\author[label3]{Tim Remiger}
\author[label3]{Michael Teutsch}

\cortext[cor1]{Corresponding Author}

\affiliation[label1]{organization={Fraunhofer Institute of Optronics, System Technologies and Image Exploitation - IOSB},
           addressline={\mbox{Gutleuthausstraße 1}}, 
           city={Ettlingen},
           postcode={76275}, 
           country={Germany}}
					
\affiliation[label2]{organization={Karlsruhe Institute of Technology - KIT},
           addressline={Kaiserstraße 12}, 
           city={Karlsruhe},
           postcode={76131}, 
           country={Germany}}
					
\affiliation[label3]{organization={HENSOLDT Sensors GmbH},
           addressline={Wörthstraße 85}, 
           city={Ulm},
           postcode={89077}, 
           country={Germany}}

\begin{abstract}
Ship detection in Synthetic Aperture Radar (SAR) images plays an important role for maritime situational awareness, especially with respect to different illegal activities at sea such as illegal fishing, smuggling or border violations. Modern ship detection methods using neural networks usually require large training datasets, which are considerably scarcer in the SAR domain than in the electro-optical domain. While several free datasets exist for this task, their availability and usability vary. In this paper, \textbf{O}pen\textbf{S}ARShip-\textbf{S}hip \textbf{D}etection \textbf{D}ataset (OSSDD), a new dataset based on the well-known OpenSARShip 1.0 dataset is proposed for training neural networks for SAR ship detection. OSSDD is freely available and contains 15,197 Sentinel-1 amplitude patches in VV and VH polarization, binary ship masks, axis-aligned bounding box and rotated bounding box annotations for a total of 55,759 ships. In the paper, the construction of the dataset, the contents and structure of the downloadable data and some experiments with three common detector models (Faster R-CNN, FCOS, DETR) are shown and discussed. The results of the detector experiments are also given to serve as benchmarks for future experiments. The dataset is available on Hugging Face: \href{https://huggingface.co/datasets/sylviaHoch/OpenSARShip-Ship-Detection-Dataset}{sylviaHoch/OpenSARShip-Ship-Detection-Dataset}.
\end{abstract}




\end{frontmatter}



\section{Introduction}

Ship detection in remote sensing data is an important task for many applications such as border surveillance, environmental monitoring purposes, oil spill detection and the detection of illegal activities as e.g. illegal fishing or smuggling. Due to its all-weather capabilities and the independence of solar illumination, SAR sensors are uniquely capable of providing continuous monitoring of maritime areas of interest. Even though many means of ship identification such as the Automatic Identification System (AIS) have been developed over the years, these systems are rather easy to bypass, and the most interesting ships are those that do not use AIS or similar systems (dark ships). Thus, (dark) ship detection in SAR images remains an important task (see e.g. [\cite{Prasad23}], [\cite{Xv320}]), even more so as SAR data have become more widely available with missions such as Sentinel-1 and the large variety of operational New Space SAR constellations. See also [\cite{Alex24}] for a very detailed review of ship detection methods and available datasets. Most modern detection algorithms are based on deep learning, i.e. neural networks, because these show superior performance to classical algorithms for many tasks. Yet, these networks require large datasets for training, which to this day are not as widely available for SAR as they are for electro-optical remote sensing. In the case of ship detection, several SAR datasets exist, which are described in Section 2 of this paper. However, most of these only contain 8-bit representations of the original SAR images, which makes their usefulness for unseen data questionable, especially if the quantization of the original SAR images to their 8-bit representation is not described. 

In this paper, we propose a new dataset named \textbf{O}pen\textbf{S}ARShip-\textbf{S}hip \textbf{D}etection \textbf{D}ataset (OSSDD) for ship detection in Sentinel-1 images that is based on the well-known OpenSARShip 1.0 dataset [\citet{Huang18}] (in the following only named OpenSARShip unless otherwise specified). While OpenSARShip was originally designed for ship classification, providing labeled image chips of individual vessels, OSSDD extends its utility to ship detection by processing the underlying Sentinel-1 scenes and leveraging the existing annotations to localize image patches containing ships. Besides the VV and VH polarized image patches in float format, the dataset comprises binary ship mask images, axis-aligned bounding boxes and rotated bounding boxes for the annotated ships, making it usable for a wide range of modern detector architectures. The dataset is freely available from \href{https://huggingface.co/datasets/sylviaHoch/OpenSARShip-Ship-Detection-Dataset}{sylviaHoch/OpenSARShip-Ship-Detection-Dataset}.

The paper is structured as follows: Section 2 contains a description of existing datasets for SAR ship detection, including links to the corresponding papers and download links to the datasets. Since OSSDD is based on OpenSARShip, Section 3 contains a brief description of this dataset. Section 4 provides a description of the processing steps that were carried out to generate the annotated images needed for OSSDD. Section 5 is dedicated to the final OSSDD dataset, a detailed description of the data and annotation formats as well as suggestions for its use and the proposed split of the dataset into train, test and validation data. Section 6 includes some experiments with different common detector architectures that were conducted by the authors to show that training such architectures with OSSDD is possible. The section also contains evaluation metrics for the best-performing detectors obtained with the training and validation parts of OSSDD on the unseen test data. These metrics can serve as a benchmark for future experiments using OSSDD. Section 7 contains the conclusions. 

\section{Existing datasets for SAR ship detection}
\label{sec1}

Since ship detection in SAR images has drawn much attention in recent years, some effort has been made by several groups to create datasets for training deep learning models for this task. The following is a list of available ship detection datasets. This list is not meant to be complete but reflects the datasets that are known to the authors at the time of writing of this paper and could be obtained without restrictions. For each of these datasets, the main properties and links to the corresponding publications are provided in Table \ref{tab:datasets}. The original SAR image sources are abbreviated as follows: Gaofen-3 (GF-3), Sentinel-1 (S-1), TerraSAR-X (TSX), TanDEM-X (TDX), Hisea-1 (HS-1), Radarsat-2 (RS-2) Umbra (Ub), Capella (Cp), Iceye (Ie). The term hp-float refers to half-precision floating point numbers. In the References section, download links to the datasets are provided under the given papers.

\begin{table}[h]
\centering
\resizebox{\textwidth}{!} {
\begin{tabular}{|l | c | c | l | l | l |}
\hline
Name & Objects & Resolution (m) & Sources & Format & Publication \\
\hline
  AIR-SARShip-1.0 & 461 & 1-3 & GF-3 & 8-bit TIF & [\citet{Sun19}]\\ 
	\hline
  HRSID & 16,951 & 0.5-3 & S-1, TSX, TDX & 8-bit PNG/JPG & [\citet{Wei20}] \\
	\hline
  SAR-Ship-Dataset & 39,729 & 3-25 & GF-3, S-1 & 8-bit JPG & [\citet{Wang19}] \\
	\hline
	SSDD & 2,456 & 1-15 & S-1, TSX, RS-2 & 8-bit JPG & [\citet{Zhang21}] \\
	\hline
	SRSDD & 2,884 & 1 & GF-3 & 8-bit PNG & [\citet{Lei21}] \\
	\hline
	OpenSARShip 1.0 & 11,346 & 2.7-17 & S-1 & float TIF& [\citet{Huang18}] \\
	\hline
	OpenSARShip 2.0 & 34,528 & 2.7-17 & S-1 & float TIF& [\citet{Li17}] \\
	\hline
	FUSARShip & 10,125 & 1 & GF-3 & 8-bit TIF & [\citet{Hou20}] \\
	\hline
	DSSDD & 3,540 & 2.3-14 & S-1 & 8-bit PNG, 16-bit TIF & [\citet{Hu21}]\\
	\hline
	xView3 & 243,018 & 10 & S-1 & hp-float TIF& [\citet{Xv320}] \\
	\hline	
	OSSDD (ours) & 55,759 & 10 & S-1 & float TIF& \\
	\hline
\end{tabular}
}
\caption{Datasets for SAR ship detection.}\label{tab:datasets}
\end{table}

Table \ref{tab:datasets} shows that numerous datasets exist for SAR ship detection. Several more exist, but are not easily accessible. Some, as those shown in Table \ref{tab:datasets}, are freely available via GitHub or Google Drive, whereas others require registration through platforms such as Baidu Cloud or radars.ac.cn, which limits their accessibility outside of China. Among the datasets listed in Table \ref{tab:datasets}, a common issue is that most are provided in 8-bit format, some even as JPG files - with the notable exception of OpenSARShip (1.0 and 2.0), DSSDD and the xView3 challenge dataset. This in itself is not problematic for training a neural network for ship detection with these datasets. However, problems may arise when applying the trained network to unseen SAR data. Since SAR amplitude data natively are recorded in float or half-precision float data formats, or as unsigned 16-bit data, the original SAR data need to be converted to 8-bit representation. There are several ways of performing this conversion, and none of the cited papers elaborate on how their data were converted to 8-bit. Thus, it is not clear how SAR data outside of the dataset should be converted to conform with the data in the dataset, which might lead to inferior performance of the neural network. Furthermore, if the images are stored as JPG, this most likely introduces lossy compression to the 8-bit data, which might further deteriorate the performance of neural networks trained with this data on unseen original SAR images.

Of the datasets shown in Table \ref{tab:datasets}, only OpenSARShip 1.0 and 2.0, DSSDD, which is rather small, and the xView3 challenge dataset do not pose the mentioned problems. OpenSARShip natively is a ship classification dataset and not primarily designed for ship detection. However, since OSSDD is based on OpenSARShip, this dataset will be described in more detail in the next section.

\section{OpenSARShip (1.0)}

OpenSARShip [\citet{Huang18}] is a large-scale open dataset for SAR ship classification. It contains the signatures of 11,346 ships extracted from 41 Sentinel-1 images of the harbor areas and their maritime surroundings of Shanghai Port (China), Shenzhen Port (China), Tianjin Port (China),
Yokohama Port (Japan), and Singapore Port (Singapore). The annotated ships are associated with AIS signals and Maritime Information System data and classified into 17 different types according to the AIS data. OpenSARShip mainly aims at ship classification, thus great care was taken that each of the provided image patches contains only one ship, centered within the patch. This leads to heterogeneous patch sizes since for most of the annotated ships, other vessels are in the surroundings, which is detrimental to ship classification. 

As this paper aims to construct a ship detection dataset, the main value of OpenSARShip lies in the fact that each image patch is linked to its source Sentinel-1 scene. As the original scenes are freely available through the Copernicus Data Space Ecosystem [\citet{Copernicus20}], OpenSARShip effectively serves as a basis for assembling a detection dataset. OpenSARShip contains 19 Single-Look Complex (SLC) images and 22 Ground-Range Detected (GRD) products with a wide range of incidence angles. The main image properties of the Sentinel-1 data in OpenSARShip are shown in Table \ref{tab:oss} in accordance with [\citet{Huang18}]. In the table, range direction is abbreviated by R, azimuth direction is abbreviated by A. OpenSARShip provides all patches in VV and VH polarization, in uncalibrated and calibrated form, and as 8-bit representation for visualization purposes.

\begin{table}[htbp]
\centering
\begin{tabular}{|l | c | c | c|}
\hline
 & Resolution R $\times$ A (m) & Looks R $\times$ A & Pixel size R $\times$ A (m) \\
\hline
GRD & 20 $\times$ 22 & 5 $\times$ 1 & 10 $\times$ 10 \\
\hline
SLC & 2.7 $\times$ 22 to 3.5 $\times$ 22 & 1 $\times$ 1 & 2.3 $\times$ 17.4\\
	\hline	
\end{tabular}
\caption{Properties of Sentinel-1 data in OpenSARShip.}\label{tab:oss}
\end{table}

\section{Construction of OSSDD}

The main idea for the construction of OSSDD was to use the same Sentinel-1 images as OpenSARShip, which already provide a wealth of different ship types and configurations. The OpenSARShip annotations were used to locate patches containing ships. To ensure consistency within OSSDD, it was decided that it should contain only GRD images. Since OpenSARShip includes both GRD and SLC images, the GRD products corresponding to the SLC images were downloaded from the Copernicus Data Space Ecosystem. No pre-processing was applied to this data. Since, during the construction of OSSDD, the original GRD data included in OpenSARShip was processed differently from the newly downloaded GRD data, the two processing pipelines are described separately in the following:

\subsection{Processing of GRD data}

For the GRD data already contained in OpenSARShip, which will be named "old GRD data" in the following, the location of many ships is known through the metadata. However, only those ships for which AIS signals were available during the construction of OpenSARShip and could be positively associated with the ship are annotated. 

\begin{figure}[h]
\centering
\begin{tabular}{c c}
\includegraphics[width = 0.25 \textwidth]{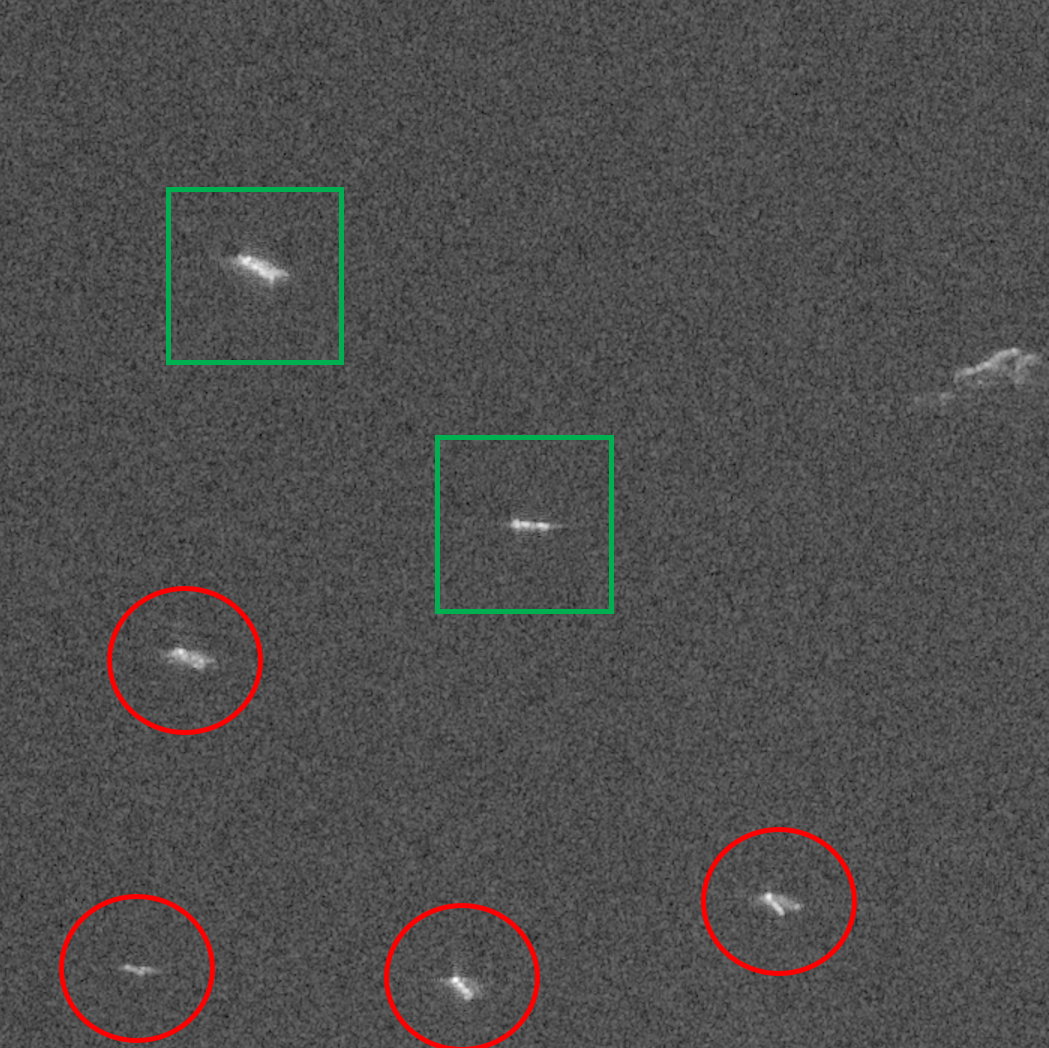} & \includegraphics[width = 0.25 \textwidth]{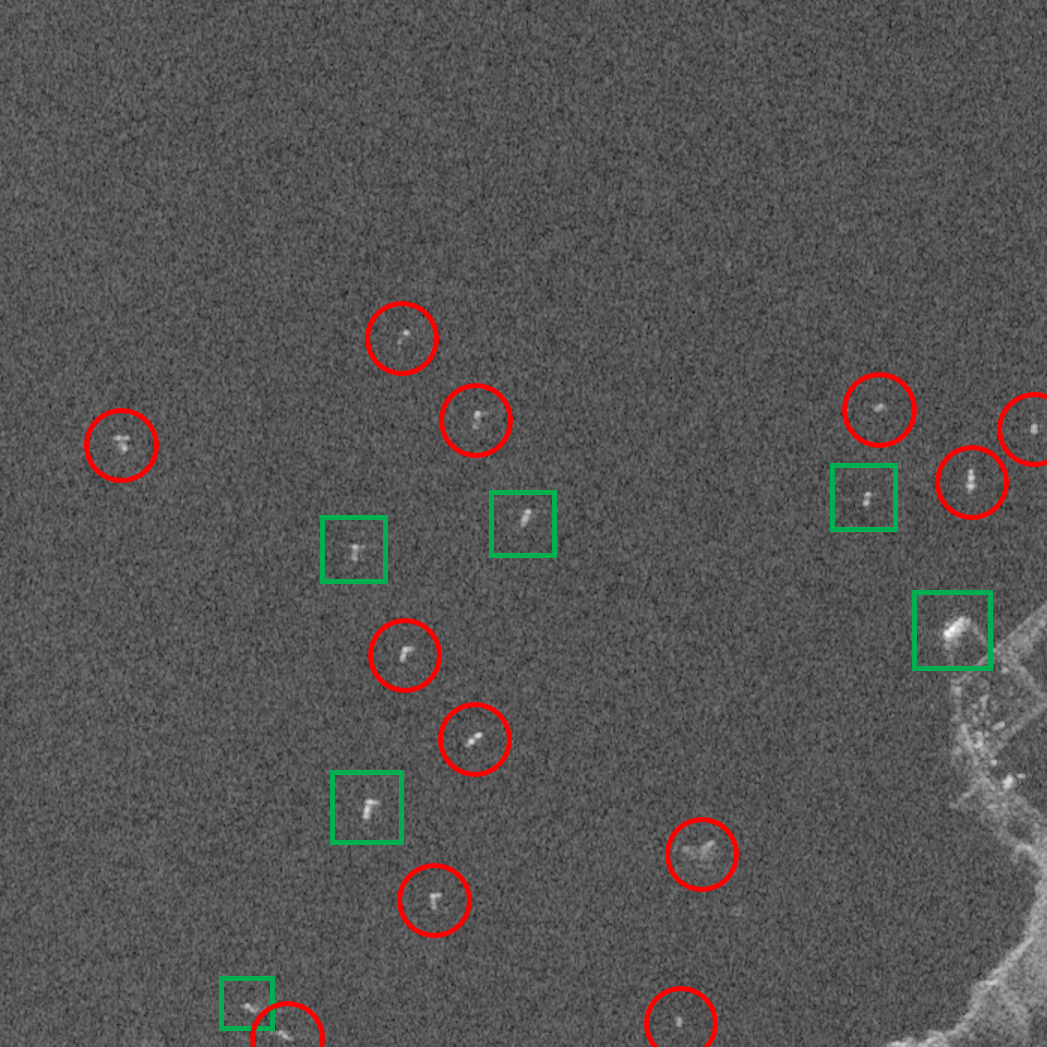} 
\end{tabular}
\caption{Sample scenes with annotated (green boxes) and missing (red circles) ships in OpenSARShip.}\label{fig:annotated}
\end{figure}

For the old GRD data, the OpenSARShip annotations were used to extract patches of size \mbox{700 $\times$ 700} pixels from the original Sentinel-1 images, centered at the annotated ships from OpenSARShip. Since the Sentinel-1 images contained in OpenSARShip depict the coastal areas of large commercial ports, most of the patches contain several ships, of which many are not annotated in OpenSARShip. Figure \ref{fig:annotated} shows example scenes, where the ships annotated in OpenSARShip are marked with a green rectangle, and those that are not annotated are marked with a red circle. It is essential to also annotate these additional ships to obtain a valid and concise ship detection dataset. For the semi-automatic annotation of additional ships, images in VH polarization are used, since in general VH polarization provides a better contrast between the ocean surface and the ships than VV polarization. 

For the extraction of the ship signatures, a heuristic threshold was applied to the Sentinel-1 images. Connected components were then extracted from the resulting binary image and very small or very narrow (1 pixel wide) components were removed, since these would pose problems with the extraction of bounding boxes later. Furthermore, some actual ships that show very large backscatterers with associated point spread functions were also excluded, since these would result in unnaturally large bounding boxes. 

\begin{figure}[h]
\centering
\begin{tabular}{c c c}
\includegraphics[width = 0.16 \textwidth]{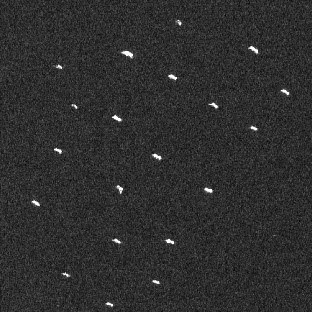} & \includegraphics[width = 0.16 \textwidth]{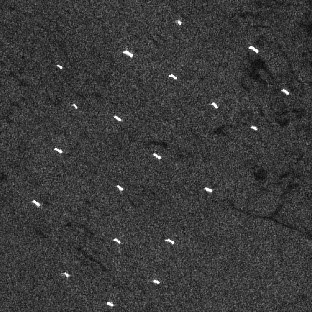} & \includegraphics[width = 0.16 \textwidth]{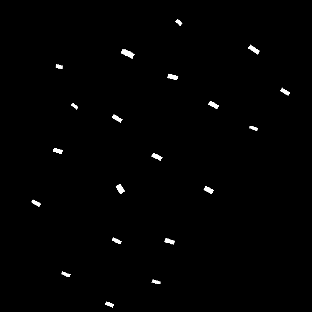} \\
\includegraphics[width = 0.16 \textwidth]{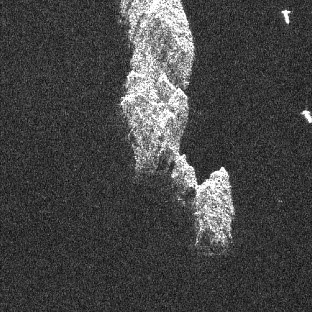} & \includegraphics[width = 0.16 \textwidth]{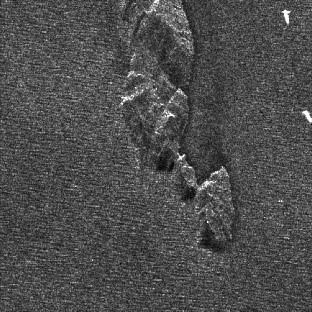} & \includegraphics[width = 0.16 \textwidth]{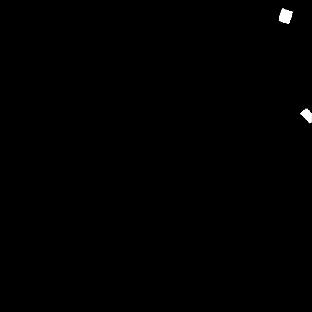} \\
\includegraphics[width = 0.16 \textwidth]{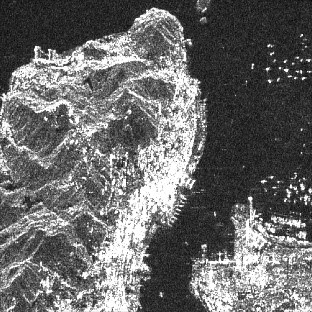} & \includegraphics[width = 0.16 \textwidth]{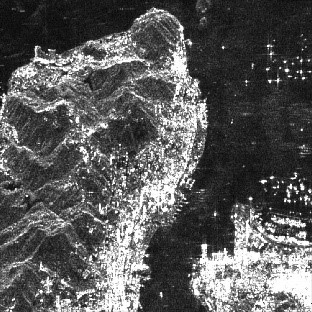} & \includegraphics[width = 0.16 \textwidth]{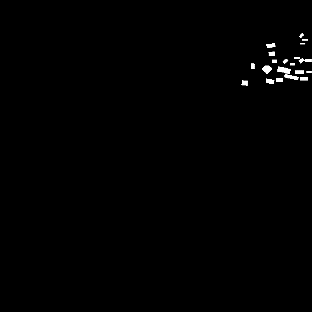} 
\end{tabular}
\caption{Examples of ship mask extractions: Left column: VH polarized Sentinel-1 scene; middle column: VV polarized Sentinel-1 scene; right column: extracted ship mask (rotated bounding boxes).}\label{fig:extr}
\end{figure}

This procedure works very well for the extraction of larger ship signatures on open water. Problems arise for smaller ships that are close together, since for these the connected components may merge and thus not represent the single ship signatures well. Image patches that show land or man-made structures may also pose problems. Due to these limitations, manual inspection and post-processing were essential to identify and remove erroneous connected components from the binary images. For the remaining connected components, axis-aligned bounding boxes (AABB) and rotated bounding boxes (RBB) were calculated. While the AABB are straightforward to calculate from the connected components, for the calculation of RBBs the algorithm provided by [\citet{Diener26}] was used. The final binary mask images were then calculated using the RBBs, marking all pixels within the RBBs as one, and all pixels outside of the RBBs as zero.

Figure \ref{fig:extr} shows some examples of extracted ship masks. It can be seen from the figure that, as expected, the VH polarized images provide much smoother and darker water surfaces than the VV polarized images. The first two rows show that the extraction of larger ships on open water works very well with the proposed algorithm. The third row shows some of the limitations of this approach. On the right side of the images there are two groups of small ships that are very close together. For the lower group, the extraction did not work well, since the signatures of different ships merged. Thus, the lower group was excluded from the binary mask. Furthermore, some smaller bright signatures were not annotated as ships since they could not be clearly identified as such.

\subsection{Processing of SLC data}
\label{subsec1}

The processing of the SLC images contained in OpenSARShip proved to be more difficult than the processing of the GRD images. As mentioned before, the goal is to provide a consistent dataset, thus the GRD version of these images was needed. The SLC images in OpenSARShip could not be converted to GRD images that exactly reproduce the properties of the GRD products provided by ESA, making the OpenSARShip annotations incompatible with the downloaded GRD images. Therefore, the annotations from OpenSARShip were not used at all. Instead, only the information about which Sentinel-1 scenes contain ships or harbor areas was leveraged. The corresponding GRD images were directly downloaded from the Copernicus Data Space Ecosystem and fully re-annotated using a semi-automatic process.

\section{The OSSDD dataset}
\label{secossdd}

The previous section described the processing of the 41 Sentinel-1 SAR GRD images corresponding to the images in OpenSARShip and the generation of the ship masks and AABB and RBB annotations. The overall raw dataset extracted from the original Sentinel-1 images consists of 15,161 image patches of size 700 $\times$ 700 pixels, for both VV and VH polarization. In these patches, a total of 64,640 ships are annotated. The data is split into training ($\sim$70\%), validation ($\sim$10\%), and test ($\sim$20\%) subsets, ensuring that patches from the same Sentinel-1 scene belong to exactly one subset and that the proportion of patches containing land is approximately balanced across splits. For the determination whether a patch contains land areas or consists exclusively of water, the so called fractional water mask computed by SNAP on the original Sentinel-1 images was used. The training subset is intended for model training, with arbitrary augmentation strategies applicable, the validation subset should be used for hyperparameter tuning, after which the best-performing model should be evaluated on the test subset.\\

The training subset comprises 10,666 patches alongside their corresponding annotations, covering a total of 37,236 annotated ships. The patch size of 700 $\times$ 700  pixels is intentionally chosen larger than typical model input sizes (e.g., 256 or 512 pixels) to facilitate data augmentation via random cropping during training, which ensures that ship signatures are not systematically centered within the input patches. For the validation and test subsets, the original patches were reduced to a size of 512 $\times$ 512 pixels via center cropping.  It should be noted that cropping operations require the metadata to be adjusted accordingly. Ships that are completely outside of the cropped area are simply dropped from the metadata, in the case of OSSDD reducing the number of annotated ships in the final dataset to 55,759. However, special care must be taken regarding ship signatures appearing at patch borders. While this does not affect the binary masks or AABB annotations, which are simply cropped to the new image dimensions, RBB annotations will no longer be rectangles if one or more vertices fall outside the cropped area. To maintain consistency and simplicity of the RBB annotations, all RBBs with at least one vertex outside the cropped image boundaries were discarded. Consequently, the annotations for binary masks and AABBs are not fully consistent with those for RBBs. The validation subset contains 1,495 patches per polarization, with 2,290 annotated ships for the binary masks and AABBs, and 2,040 annotated ships for the RBBs. The test subset consists of 3,036 patches per polarization, comprising 15,573 ship annotations for the binary masks and AABBs, and 14,145 for the RBBs. A summary of all dataset statistics is provided in Table~\ref{tab:prop}.

\begin{table}[h]
\centering
\begin{tabular}{|l | c | c | c| c|}
\hline
 & Patch size (px) & No. patches & No. obj. Bin and AABB & No. obj. RBB \\
\hline
Train & 700 $\times$ 700 & 10,666 & 37,236 & 37,236 \\
\hline
Test & 512 $\times$ 512 & 3,036 & 15,573 & 14,145\\
	\hline	
Validate & 512 $ \times$ 512 & 1,495 & 2,990 & 2,040 \\
\hline
Total &  & 15,197 & 55,759 & 53,421 \\
\hline
\end{tabular}
\caption{Summary of OSSDD statistics.}\label{tab:prop}
\end{table}

\subsection{Structure of the dataset}

The dataset comprises five directories:

\begin{itemize}
\item \textbf{Chip\_VV} contains the cutout patches in VV polarization as 32-bit TIF files.
\item \textbf{Chip\_VH} contains the cutout patches in VH polarization as 32-bit TIF files.
\item \textbf{Binary\_Masks} contains the binary ship masks as 8-bit TIF files.
\item \textbf{Metadata\_AABB} contains the annotated axis-aligned bounding box coordinates as ASCII TXT files.
\item \textbf{Metadata\_RBB} contains the annotated rotated bounding box coordinates as ASCII TXT files.
\item Additionally, a CSV file (metadata.csv) is given that specifies the land cover percentage for each patch and indicates whether it belongs to the training, test, or validation set.
\end{itemize}

\subsection{Metadata for AABB}

The metadata files for the AABBs are located in a separate folder named "Metadata\_AABB". Each of the ASCII text files in this folder has the structure shown in Table \ref{tab:aabb}: 

\begin{table}[h]
\centering
\begin{tabular}{l l l l l l l}
object\_count & & & & & & \\
object\_id & min\_x & min\_y & max\_x & max\_y & center\_x & center\_y \\
\end{tabular}
\caption{Structure of the metadata files for AABB.}\label{tab:aabb}
\end{table}

Here, object\_count is the number of extracted bounding boxes in the patch. The following lines provide the object id of the bounding box, followed by the corner coordinates of the AABB. The last two entries in each line are the center coordinates of the AABB, rounded to integer. These coordinates are valid for image patches with coordinate (0,0) at the top left corner of the image and x-axis going from left to right and y-axis going from top to bottom of the images.

\subsection{Metadata for RBB}

The metadata files for the RBBs are located in a separate folder named "Metadata\_RBB". Each of the ASCII text files in this folder has the structure shown in Table \ref{tab:rbb}: 

\begin{table}[h]
\centering
\begin{tabular}{l l l l l l l l l l l}
object\_count & & & & & & & & & &\\
object\_id & x$_1$ & y$_1$ & x$_2$ & y$_2$ & x$_3$ & y$_3$ & x$_4$ & y$_4$ & center\_x & center\_y \\
\end{tabular}
\caption{Structure of the metadata files for RBB.}\label{tab:rbb}
\end{table}

The first integer object\_count is again the number of bounding boxes in the image patch. The following lines provide the object id of the bounding box, followed by the corner coordinates of the four corner points of the RBB. As for the AABB, the last two entries in each line are the center coordinates of the RBB, rounded to integer. These coordinates again are valid for image patches with coordinate (0,0) at the top left corner of the image and x-axis running from left to right and y-axis running from top to bottom of the image.

\section{Ship detection experiments}
\label{exp}

Several ship detection experiments with different detector network architectures were carried out, which are presented in this section. The results obtained on the unseen test data of OSSDD are also presented. These are meant to be used as a baseline for future ship detection experiments using OSSDD.

\subsection{Architectures chosen for the ship detection experiments}

OSSDD was tested on several common detection frameworks with different properties. The chosen networks were Faster R-CNN [\citet{Ren15}], Fully Convolutional One-Stage Object Detection (FCOS) [\citet{Tian19}] and the transformer-based detector Detection Transformer (DETR) [\citet{Carion20}]. These represent different paradigms for object detection: 

Faster R-CNN is a two-stage anchor-free detector, for which in the first stage a Region Proposal Network (RPN) generates a set of anchor candidates, which in the second stage are classified and refined by bounding box regression. This makes the network depend heavily on the choice and number of anchor boxes but the explicit separation of the generation of candidates and object detection usually gives reliable results.

In contrast to Faster R-CNN, FCOS is a one-stage anchor-free detector which predicts class, object center and bounding box offsets for each position on the feature map directly. Thus, no RPN, no pre-defined anchor boxes and no anchor hyperparameters are needed, simplifying the architecture.

DETR defines object detection as a set matching problem and uses a transformer decoder with a fixed set of object queries to identify and localize a set of objects. In contrast to Faster R-CNN and FCOS, DETR uses no heuristics such as Non-Maximum-Suppression, anchors or RPN and rather learns global context relationships using self- and cross-attention mechanisms.

A ResNet-50 model with Feature Pyramid Network was used as backbone for image feature extraction for all detectors. For the training of Faster R-CNN and FCOS, both the backbone and the detection head were initialized with random weights, and both detectors were trained from scratch. For DETR, the training was based on a pre-trained model [\citet{Carion20}], since convergence could not be achieved for this detector when trained from scratch.

\subsection{Detector training}

All three detectors were trained separately for the VV and the VH data using the Adam optimizer [\cite{kingma17}] with decoupled weight decay and the following parameters: $\beta_1=0.9$, $\beta_2=0.999,$ $\textup{weight\_decay}=0,$ $\epsilon_{\textup{adam}}=10^{-8}$ and the initial learning rate was set to $\gamma=10^{-4}$, which was lowered during the training using a cosine annealing scheduler. All three detectors used the training part of OSSDD described in Section 5 for training and the validation part for hyperparameter tuning. Prior to training, the SAR amplitude values were converted to decibel scale to reduce the high dynamic range, and each patch was subsequently normalized to the range [0, 1] via linear min-max scaling based on the per-patch minimum and maximum values. Furthermore, on-the-fly data augmentation was performed for the training data by randomly cropping patches of size 512 × 512 pixels from the original 700 $\times$ 700 pixel patches at the beginning of each epoch. To account for the imbalance between complex and simple samples, two random crops were extracted from patches with a land fraction exceeding 5\%, while a single crop was extracted from the remaining patches consisting almost exclusively of water background. The test data served as holdout dataset to evaluate the performance of the final trained models on unseen data. All models were trained for 25 epochs using a batch size of 8 and the best achieved result, evaluated on the validation set, was saved and then used to evaluate the model performance on the unseen test dataset.

\subsection{Performance evaluation}

Performance of the trained models was evaluated using precision, recall and mean Average Precision (mAP). Precision encodes the proportion of correctly detected objects among all detected objects, while recall encodes the proportion of correctly identified objects among all truly present objects. A threshold of 0.2 for the Intersection over Union (IoU) of the detected bounding boxes with the true bounding boxes was used along with a confidence threshold of 0.3 for the calculation of precision and recall. mAP measures the area under the precision-recall curve for several IoU thresholds and thus is an aggregated measure of detection performance. In this paper, mAP was calculated over IoU thresholds between 0.1 and 0.95 in steps of 0.05.

The results for the different detectors on the VH test set of OSSDD are shown in Table \ref{tab:erg}, those for the VV polarized test set of OSSDD are contained in Table \ref{tab:erg2}.

\begin{table}[h]
\centering
\begin{tabular}{ | l | c | c | c|}
\hline
Detector & Precision & Recall & mAP \\
\hline
Faster R-CNN & 0.7442 & 0.8942 & 0.7018 \\
\hline
FCOS & 0.8424 & 0.8091 & 0.6961\\
\hline
DETR & 0.7444 & 0.8891 & 0.5999 \\
	\hline	
\end{tabular}
\caption{Results for the three detectors on the VH test set of OSSDD.}\label{tab:erg}
\end{table}

\begin{table}[h]
\centering
\begin{tabular}{| l | c | c | c|}
\hline
Detector & Precision & Recall & mAP \\
\hline
Faster R-CNN & 0.5981 & 0.8808 & 0.6228 \\
\hline
FCOS & 0.8153 & 0.8056 & 0.6210\\
\hline
DETR & 0.6954 & 0.8789 & 0.5220 \\
	\hline	
\end{tabular}
\caption{Results for the three detectors on the VV test set of OSSDD.}\label{tab:erg2}
\end{table}

\begin{figure}[h]
\centering
\begin{tabular}{c c }
\includegraphics[width = 0.23 \textwidth]{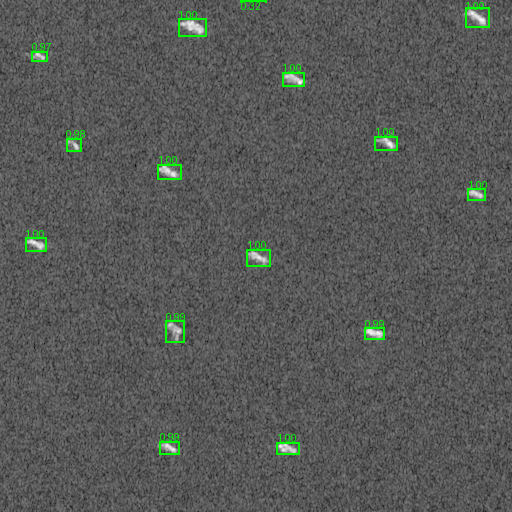} a) & \includegraphics[width = 0.23 \textwidth]{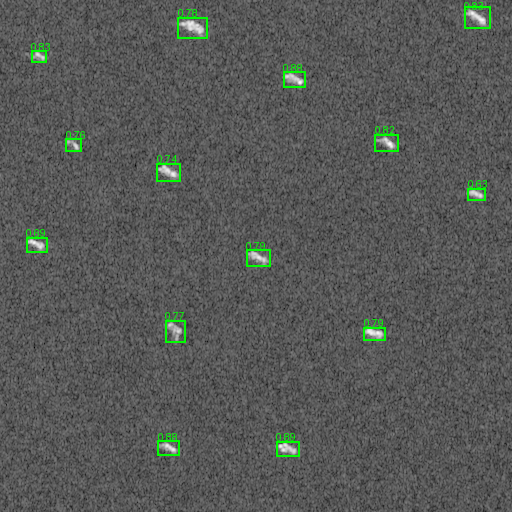} b) \\
\includegraphics[width = 0.23 \textwidth]{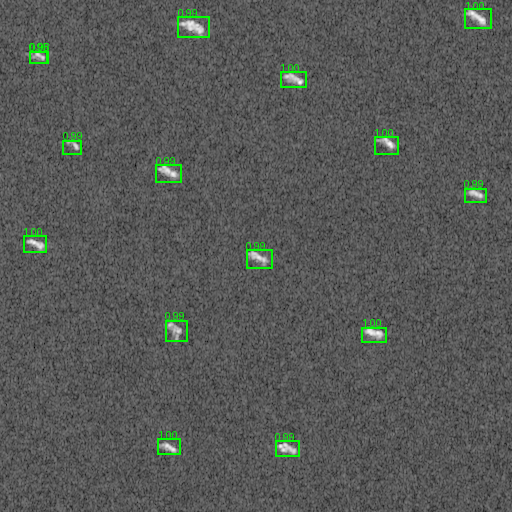} c) & \includegraphics[width = 0.23 \textwidth]{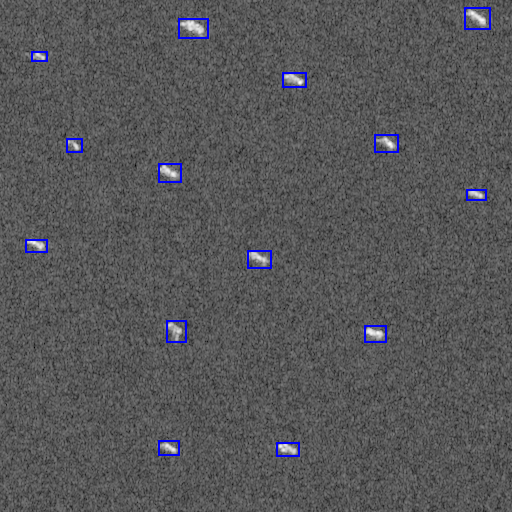} d)
\end{tabular}
\caption{Detection results for ships on open water. a) Result of Faster R-CNN; b) Result of FCOS; c) Result of DETR; d) Ground truth.}\label{fig:result1}
\end{figure}

For a more detailed analysis of the detection performance, in the next figures some detection results with good performance and some of the more problematic image patches are shown: Figure \ref{fig:result1} shows that all three detectors perform well in scenarios with several ships that are well separated and in open water. All of the ships were detected and the predicted bounding boxes align well with the ground truth.

\begin{figure}[h!]
\centering
\begin{tabular}{c c}
\includegraphics[width = 0.23 \textwidth]{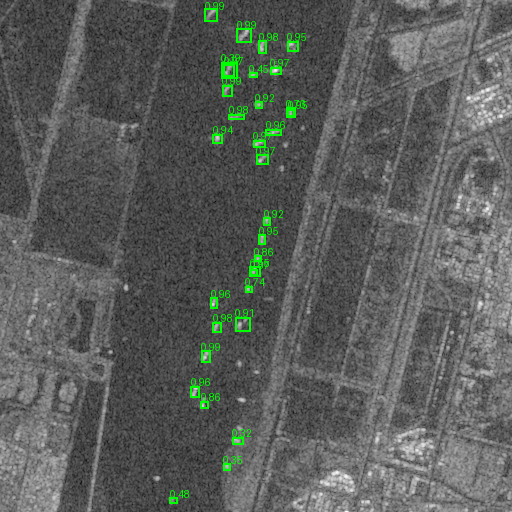} a) & \includegraphics[width = 0.23 \textwidth]{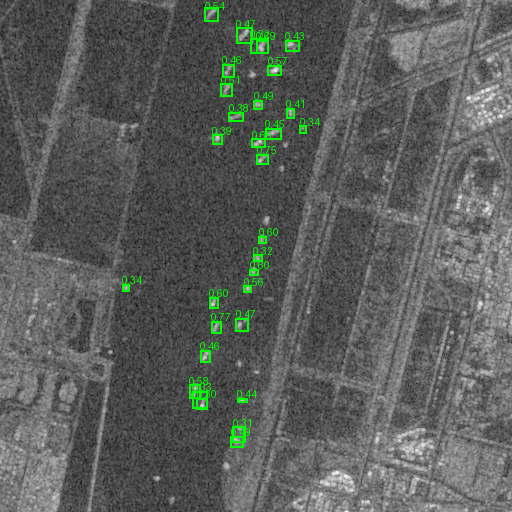} b)\\
\includegraphics[width = 0.23 \textwidth]{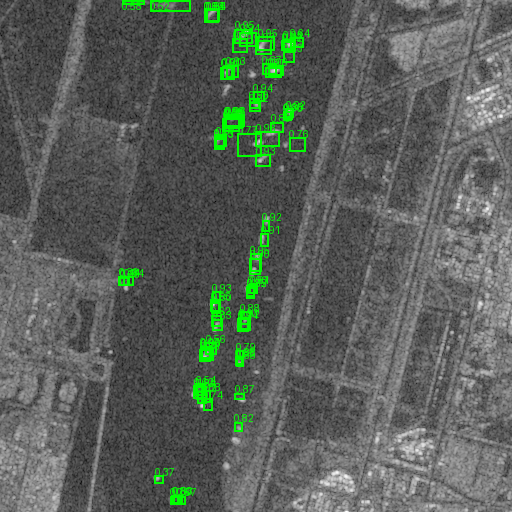} c) & \includegraphics[width = 0.23 \textwidth]{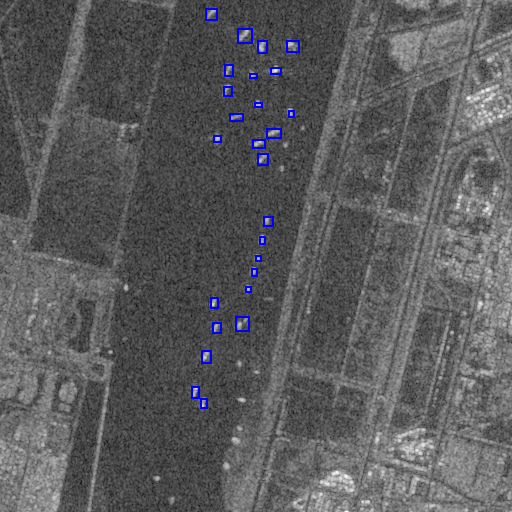} d)
\end{tabular}
\caption{Detection results for smaller ships close together and close to land structures. a) Result of Faster R-CNN; b) Result of FCOS; c) Result of DETR; d)  Ground truth.}\label{fig:result2}
\end{figure}

Figure \ref{fig:result2} shows a more difficult scenario with many small ships located close to land structures. Here, all of the detectors struggle to detect all of the annotated ships, and also detect some brighter land structures that were not annotated as ships. Remarkably, DETR shows multiple detections for most of the ships, which is not the case for Faster R-CNN and FCOS.

 \begin{figure}[h]
\centering
\begin{tabular}{c c}
\includegraphics[width = 0.23 \textwidth]{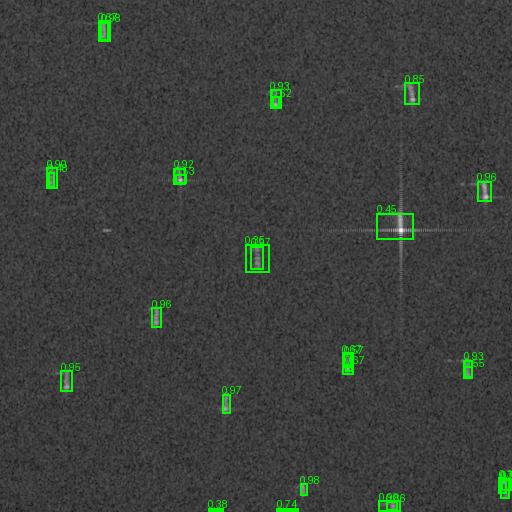} a) & \includegraphics[width = 0.23 \textwidth]{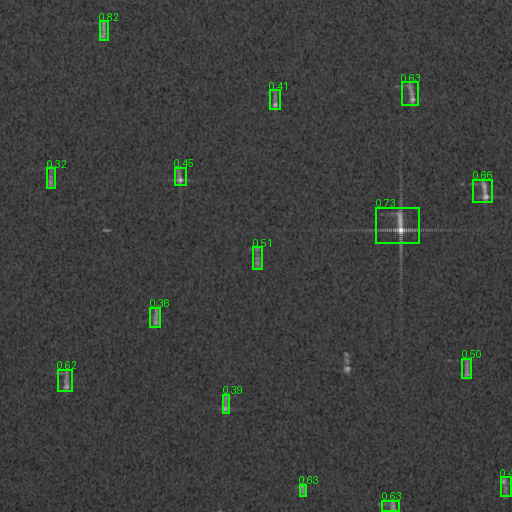} b)\\
\includegraphics[width = 0.23 \textwidth]{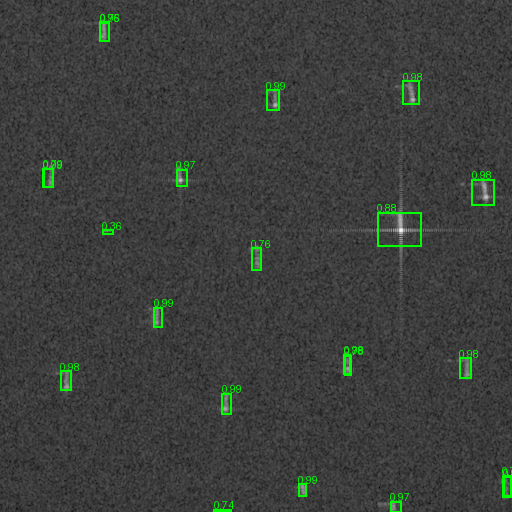} c) & \includegraphics[width = 0.23 \textwidth]{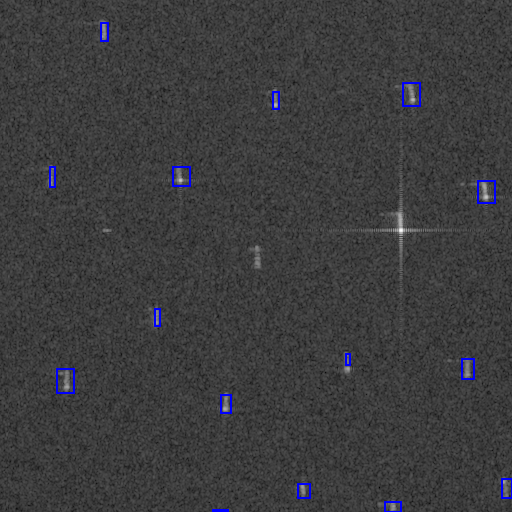} d)
\end{tabular}
\caption{Detection results for difficult ships. a) Result of Faster R-CNN; b) Result of FCOS; c) Result of DETR; d) Ground truth.}\label{fig:result3}
\end{figure}

Figure \ref{fig:result3} shows some of the problems one is faced with in the creation of a SAR ship detection dataset and in the evaluation of the detection results. The ship located at the right side of the image showing a very bright scattering center was excluded from the ground truth data because the very bright scatterer would lead to very large bounding boxes which may mislead the training. Yet, all three detectors were able to find the ship and provide it with a reasonable bounding box. This will however be classified as a misdetection. The same is true for the ship in the center of the image whose signature is rather weak and thus was excluded from the ground truth. The DETR result also shows a true misdetection at the right side of the image towards the middle, where actually a side lobe of the very bright scatterer was detected as a ship.\\

In general, aside from the mentioned problems, Figures \ref{fig:result1} - \ref{fig:result3} show that all of the detectors work rather well when trained with OSSDD, which is also confirmed by the quantitative evaluation shown in Tables \ref{tab:erg} and \ref{tab:erg2}. 

\section{Conclusion}

In this paper a new dataset named OpenSARShip-Ship Detection Dataset (OSSDD) was introduced. It is based on OpenSARShip 1.0 and contains 15,197 image patches extracted from Sentinel-1 GRD image products, with 55,759 semi-automatically annotated ship signatures. Unlike most existing SAR ship detection datasets, which are distributed in 8-bit or lossy compressed formats, OSSDD provides images in 32-bit floating point format, ensuring full compatibility with original SAR data and avoiding potential performance degradation due to unknown quantization procedures. The image patches are included in VV and VH polarization and the annotations as binary images, Axis-Aligned Bounding Boxes and Rotated Bounding Boxes. Thus, OSSDD can be used with a wide range of detector architectures. The construction of the dataset has been described in detail, including a discussion on problematic detections and on the reasons why some detections were excluded from the annotations. Results of Sentinel-1 ship detection experiments with three neural network based detectors were shown for VH and VV polarization. The visual comparison of the results with the ground truth annotations show that all three detectors could be trained using OSSDD. The best-performing detector (Faster R-CNN) achieved a mAP of 0.70 on the VH test set, demonstrating that OSSDD provides a sufficiently rich and well-annotated training basis for detectors based on Deep Learning. OSSDD and the presented detector results are intended to serve as a reproducible benchmark for the SAR ship detection community. While the dataset provides a standardized basis for training and evaluating new detection approaches, the detector results establish concrete baseline performances against which future methods can be objectively compared.

\bibliographystyle{elsarticle-harv} 
 \bibliography{elsarticle-template-harv.bib}

@article{Alex24,
  author={Alexandre, Cyprien and Devillers, Rodolphe and Mouillot, David and Seguin, Raphael and Catry, Thibault},
  journal={IEEE Journal of Selected Topics in Applied Earth Observations and Remote Sensing}, 
  title={{Ship Detection With SAR C-Band Satellite Images: A Systematic Review}}, 
  year={2024},
  volume={17},
  number={},
  pages={pp. 14353--14367},
  doi={10.1109/JSTARS.2024.3437187}}

@inproceedings{Carion20,
author = {Carion, Nicolas and Massa, Francisco and Synnaeve, Gabriel and Usunier, Nicolas and Kirillov, Alexander and Zagoruyko, Sergey},
title = {{End-to-End Object Detection with Transformers}},
year = {2020},
publisher = {Springer-Verlag},
address = {Berlin, Heidelberg},
doi = {10.1007/978-3-030-58452-8\_13, Link DETR: \\https://github.com/facebookresearch/detr/tree/main?tab=readme-ov-file (last visited 07/07/2026)},
booktitle = {Computer Vision – ECCV 2020: 16th European Conference, Glasgow, UK, August 23–28, 2020, Proceedings, Part I},
pages = {213--229}}

@article{Hou20,
author = {Hou, Xiyue and Ao, Wei and Song, Qian and Lai, Jian and Wang, Haipeng and Xu, Feng},
year = {2020},
month = {04},
pages = {},
title = {{FUSAR-Ship: building a high-resolution SAR-AIS matchup dataset of Gaofen-3 for ship detection and recognition}},
volume = {63},
journal = {Science China Information Sciences},
doi = {10.1007/s11432-019-2772-5, Dataset link: \\https://drive.google.com/file/d/1SOEMud9oUq69gxbfcBkOvtUkZ3LWEpZJ/view (last visited 07/06/2026).}}

@ARTICLE{Hu21,
  author={Hu, Y and Li, Y and Pan, Z},
	journal={Sensors (Basel)},
	title={{A Dual-Polarimetric SAR Ship Detection Dataset and a Memory-Augmented Autoencoder-Based Detection Method}},
	year={2021},
	volume={24},
	pages={8478},
	doi={10.3390/s21248478, Dataset link: \\https://github.com/liyiniiecas/A\_Dual-polarimetric\_SAR\_Ship\_Detection\_Dataset (last visited 07/13/2026)}}

@ARTICLE{Huang18,
  author={Huang, Lanqing and Liu, Bin and Li, Boying and Guo, Weiwei and Yu, Wenhao and Zhang, Zenghui and Yu, Wenxian},
  journal={IEEE Journal of Selected Topics in Applied Earth Observations and Remote Sensing}, 
  title={{OpenSARShip: A Dataset Dedicated to Sentinel-1 Ship Interpretation}}, 
  year={2018},
  volume={11},
  number={1},
  pages={pp.~195--208},
  doi={10.1109/JSTARS.2017.2755672, Dataset link: \\https://opensar.sjtu.edu.cn/DataAndCodes.html (last visited 07/06/2026)}}

@Article{Lei21,
AUTHOR = {Lei, Songlin and Lu, Dongdong and Qiu, Xiaolan and Ding, Chibiao},
TITLE = {{SRSDD-v1.0: A High-Resolution SAR Rotation Ship Detection Dataset}},
JOURNAL = {Remote Sensing},
VOLUME = {13},
YEAR = {2021},
NUMBER = {24},
ARTICLE-NUMBER = {5104},
DOI = {10.3390/rs13245104, Dataset link: \\https://github.com/HeuristicLU/SRSDD-V1.0?tab=readme-ov-file (last visited 07/06/2026)}}

@misc{Xv320,
      title={{xView3-SAR: Detecting Dark Fishing Activity Using Synthetic Aperture Radar Imagery}}, 
      author={Fernando Paolo and T. T. Lin and Ritwik Gupta and Bryce Goodman and Nirav Patel and Daniel Kuster and David Kroodsma and Jared Dunnmon},
      year={2022},
      howpublished={https://arxiv.org/abs/2206.00897}, 
}

@inproceedings{Ren15,
author = {Ren, Shaoqing and He, Kaiming and Girshick, Ross and Sun, Jian},
title = {{Faster R-CNN: Towards real-time object detection with region proposal networks}},
year = {2015},
publisher = {MIT Press},
address = {Cambridge, MA, USA},
booktitle = {{Proceedings of the 29th International Conference on Neural Information Processing Systems - Volume~1}},
pages = {91--99},
numpages = {9},
location = {Montreal, Canada},
series = {NIPS'15}}

@article{Sun19,
title = {{AIR-SARShip-1.0: High-resolution SAR Ship Detection Dataset}},
journal = "Journal of Radars",
volume = "8",
number = "R19097",
pages = "852",
year = "2019",
doi = "10.12000/JR19097, Dataset link: \\https://radars.ac.cn/web/data/getData?newsColumnId=d25c94d7-8fe8-415f-a897-cb88657a8141\&pageType=en (last visited 07/06/2026)",
author = {Sun, Xian and Wang, Zhirui and Sun, Yuanrui and Diao, Wenhui and Zhang Yue and Fu, Kun}}

@article{Tian19,
  title={{FCOS: Fully Convolutional One-Stage Object Detection}},
  author={Zhi Tian and Chunhua Shen and Hao Chen and Tong He},
  journal={2019 IEEE/CVF International Conference on Computer Vision (ICCV)},
  year={2019},
  pages={pp.~9626--9635},
  howpublished={https://api.semanticscholar.org/CorpusID:91184137}
}

@Article{Wang19,
AUTHOR = {Wang, Yuanyuan and Wang, Chao and Zhang, Hong and Dong, Yingbo and Wei, Sisi},
TITLE = {{A SAR Dataset of Ship Detection for Deep Learning under Complex Backgrounds}},
JOURNAL = {Remote Sensing},
VOLUME = {11},
YEAR = {2019},
NUMBER = {7},
ARTICLE-NUMBER = {765},
DOI = {10.3390/rs11070765, Dataset link: \\https://github.com/CAESAR-Radi/SAR-Ship-Dataset (last visited 07/06/2026)}}

@ARTICLE{Wei20,
  author={Wei, Shunjun and Zeng, Xiangfeng and Qu, Qizhe and Wang, Mou and Su, Hao and Shi, Jun},
  journal={IEEE Access}, 
  title={{HRSID: A High-Resolution SAR Images Dataset for Ship Detection and Instance Segmentation}}, 
  year={2020},
  volume={8},
  number={},
  pages={pp. 120234--120254},
  doi={10.1109/ACCESS.2020.3005861, Dataset link: \\https://github.com/chaozhong2010/HRSID (last visited 07/06/2026)}}

@Article{Zhang21,
AUTHOR = {Zhang, Tianwen and Zhang, Xiaoling and Li, Jianwei and Xu, Xiaowo and Wang, Baoyou and Zhan, Xu and Xu, Yanqin and Ke, Xiao and Zeng, Tianjiao and Su, Hao and Ahmad, Israr and Pan, Dece and Liu, Chang and Zhou, Yue and Shi, Jun and Wei, Shunjun},
TITLE = {{SAR Ship Detection Dataset (SSDD): Official Release and Comprehensive Data Analysis}},
JOURNAL = {Remote Sensing},
VOLUME = {13},
YEAR = {2021},
NUMBER = {18},
ARTICLE-NUMBER = {3690},
DOI = {10.3390/rs13183690, Dataset link: \\https://github.com/TianwenZhang0825/Official-SSDD (last visited 07/06/2026)}}

@misc{Copernicus20,
	author = {ESA},
	title = {{Copernicus Dataspace}},
	year = {2026},
  howpublished = {https://browser.dataspace.copernicus.eu/ (last visited 07/07/2026)}}

@misc{Diener26,
	author = {Julien Diener},
	title = {{2D minimal bounding box}},
	year = {2026},
  howpublished = {Link: https://de.mathworks.com/matlabcentral/fileexchange/31126-2d-minimal-bounding-box (last visited 07/07/2026)},
  }

@INPROCEEDINGS{Prasad23,
  author={Prasad, M. S. and Verma, Shivani and Shichkina, Yulia A.},
  booktitle={2023 XXVI International Conference on Soft Computing and Measurements (SCM)}, 
  title={{Dark Ship detection: SAR Images}}, 
  year={2023},
  volume={},
  number={},
  pages={341-344},
  doi={10.1109/SCM58628.2023.10159062}}

@INPROCEEDINGS{Li17,
  author={Li, Boying and Liu, Bin and Huang, Lanqing and Guo, Weiwei and Zhang, Zenghui and Yu, Wenxian},
  booktitle={2017 SAR in Big Data Era: Models, Methods and Applications (BIGSARDATA)}, 
  title={{OpenSARShip 2.0: A large-volume dataset for deeper interpretation of ship targets in Sentinel-1 imagery}}, 
  year={2017},
  volume={},
  number={},
  pages={1--5},
  doi={10.1109/BIGSARDATA.2017.8124929, Dataset link: \\https://opensar.sjtu.edu.cn/DataAndCodes.html (last visited 07/13/2026)}}

@misc{kingma17,
      title={Adam: A Method for Stochastic Optimization}, 
      author={Diederik P. Kingma and Jimmy Ba},
      year={2017},
      howpublished={https://arxiv.org/abs/1412.6980}, 
}

\end{document}